\documentclass[11pt,a4paper]{article}
\usepackage[hyperref]{acl2020}
\usepackage{times}
\usepackage{color,colortbl}
\usepackage{latexsym}

\usepackage{graphicx,todonotes,booktabs,enumitem,multirow}
\usepackage{caption,subcaption}

\definecolor{lightgray}{rgb}{0.9, 0.9, 0.9}

\usepackage{microtype}

\aclfinalcopy 

\newcommand\BibTeX{B\textsc{ib}\TeX}

 \title{On the interaction of automatic evaluation\\ and task framing in headline style transfer}

\author{\bf {Lorenzo De Mattei}$^{\star}$$^{\diamond}$$^{\dagger}$, \bf{Michele Cafagna$^{\ddagger}$, \bf{Huiyuan  Lai}$^{\dagger}$},\\ {\bf Felice Dell'Orletta$^{\diamond}$, Malvina Nissim$^{\dagger}$, Albert Gatt$^{\ddagger}$} \\ 
\textsuperscript{$\star$} Department of Computer Science, University of Pisa / Italy \\
  \textsuperscript{$\diamond$} ItaliaNLP Lab, Istituto di Linguistica Computazionale  ``Antonio Zampolli'', Pisa / Italy \\
  \textsuperscript{$\dagger$} CLCG, University of Groningen / The Netherlands \\
  \textsuperscript{$\ddagger$} LLT, University of Malta / Malta \\
  {\tt lorenzo.demattei@di.unipi.it} \\
   {\tt \{michele.cafagna,albert.gatt\}@um.edu.mt}\\
   {\tt \{h.lai,m.nissim\}@rug.nl}\\
   {\tt felice.dellorletta@ilc.cnr.it}\\
    }

\begin{document}

\maketitle

\begin{abstract}
An ongoing debate in the NLG community concerns the best way to evaluate systems, with human evaluation often being considered the most reliable method, compared to corpus-based metrics. However, tasks involving subtle textual differences, such as style transfer, tend to be hard for humans to perform. In this paper, we propose an evaluation method for this task based on purposely-trained classifiers, showing that it better reflects system differences than traditional metrics such as BLEU and ROUGE.
\end{abstract}

\section{Introduction and Background}\label{sec:intro}

The evaluation of Natural Language Generation (NLG) systems is intrinsically complex. This is in part due to the virtually open-ended range of possible ways of expressing content, making it difficult to determine a `gold standard' or `ground truth'. As a result, there has been growing scepticism in the field surrounding the validity of corpus-based metrics, primarily because of their weak or highly variable correlations with human judgments \cite{ReiterSripada2002,Reiter2009a,Reiter2018,celikyilmaz2020evaluation}. Human evaluation is generally viewed as the most desirable method to assess generated text \cite{novikova-etal-2018-rankme,van-der-lee-etal-2019-best}. In their recent comprehensive survey on the evaluation of NLG systems, \citet{celikyilmaz2020evaluation} stress that it is important that any used untrained automatic measure (such as BLEU, ROUGE, METEOR, etc)  correlates well with human judgements.

At the same time, human evaluation also presents its challenges and there have been calls for the development of new, more reliable metrics \cite{Novikova2017}. Beyond the costs associated with using humans in the loop during development, it also appears that certain linguistic judgment tasks are hard for humans to perform reliably. For instance, human judges show relatively low agreement in the presence of syntactic variation \cite{cahill-forst-2009-human}. By the same token, \citet{dethlefs2014cluster} observe at best moderate correlations between human raters on stylistic dimensions such as politeness, colloquialism and naturalness.

Closer to the concerns of the present work, it has recently been shown that humans find it difficult to identify subtle stylistic differences between texts. \citet{invisible-dema-2020} presented three independent judges with headlines from two Italian newspapers with distinct ideological leanings and in-house editorial styles. When asked to classify the headlines according to which newspaper they thought they came from, all three annotators performed the task with low accuracy (ranging from 57\% to 62\%). Furthermore, agreement was very low (Krippendorff's $\alpha = 0.16$). Agreement was similarly low on classifying automatically generated headlines ($\alpha=0.13$ or $0.14$ for two different generation settings). These results suggest that human evaluation is not viable, or at least not sufficient, for this task. 










In this work we focus on the same 
style-transfer task using headlines from newspapers in Italian, but address the question of whether a series of classifiers that monitor both style strength as well as content preservation, the core aspects of style transfer \cite{fu2018style,mir-etal-2019-evaluating, luo-2019}, can shed light on differences between models. 

We also add some untrained automatic metrics for evaluation. As observed above, the fact that humans cannot perform this task reliably makes it impossible to choose such metrics based on good correlations with human judgement \cite{celikyilmaz2020evaluation}. Therefore, relying on previous work, we compare the insights gained from our classifiers with those obtained from BLEU \cite{papineni-etal-2002-bleu} and ROUGE \cite{lin-2004-rouge}, since they are commonly used metrics to assess performance for content preservation and summarisation. Other common metrics such as METEOR \cite{banerjee2005meteor} and BLEURT  \cite{sellam2020bleurt}, which in principle would be desirable to use,  are not applicable to our use case as they require resources not available for Italian.

More specifically, we train a classifier which, given a headline coming from one of two newspapers with distinct ideological leanings and in-house styles, can identify the provenance of the headline with high accuracy. We use this (the `main' classifier) to evaluate the success of a model in regenerating a headline from one newspaper, in the style of the other. We add two further consistency checks, both of which aim at content assessment, and are carried out using additional classifiers trained for the purpose: (a) a model's output headline should still be compatible in content with the original headline; (b) the output headline should also be compatible in content with the article to which it pertains. A headline is deemed to be (re)generated successfully in a different style if both (a) and (b) are satisfied, 
and the main classifier's decision as to its provenance should be reversed, relative to its decision on the original headline. 

A core element in our setup is testing our evaluation classifiers/strategies in different scenarios that arise from different ways of framing the style transfer task, and different degrees of data availability. Indeed, we frame the task either as a translation problem, where a headline is rewritten in the target style or as a summarisation problem, where the target headline is generated starting from the source article, using a summarisation model trained on target style. The two settings differ in their needs in terms of training data as well as in their ability to perform the two core aspects of style transfer (style strength and content preservation).  

We observe how evaluation is affected by the different settings, and how this should be taken into account when deciding what the best model is.

Data and code used for this paper are available at \url{https://github.com/michelecafagna26/CHANGE-IT}. The data and task settings also lend themselves well as material for a shared task, and they have indeed been used, with the summarisation system described here as baseline, in the context of the EVALITA~2020 campaign for Italian NLP \cite{change-it:2020}.




 


\begin{figure*}[h]
    \begin{subfigure}[c]{0.6\textwidth}
         \centering
    \includegraphics[width=.94\columnwidth]{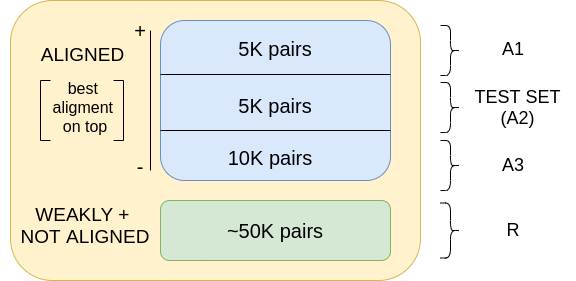}
    \caption{\label{fig:data-setting} Data splits
    }
     \end{subfigure}
     \begin{subfigure}[c]{0.4\textwidth}
         \centering
          \begin{tabular}{l|ll}\toprule
 \multicolumn{3}{c}{training sets}\\
 \midrule
\multirow{3}{*}{EVAL}    & main & R+A3+A1 \\
    &  HH& A1 + random pairs\\
    &  AH & R+A3+A1\\ \midrule
\multirow{3}{*}{TASK}    & \texttt{SUM} & R+A3\\
\cline{2-3}
    & \texttt{S2S1} & A3 (10K) \\
    & \texttt{S2S2} & A3+A1 (15K)\\
    & \texttt{S2S3} & A1 (5K) \\
\bottomrule
 \end{tabular}
         \caption{Training sets}
         \label{fig:trainsets}
     \end{subfigure}
     \caption{Data splits and their use in the different training sets}
\end{figure*}

\section{Task and Data} 
Our style transfer task 
can be seen as a ``headline translation" problem. Given a collection of headlines from two newspapers at opposite ends of the political spectrum, the task is to change all rightwing headlines to headlines with a leftwing style, and all leftwing headlines to headlines with a rightwing style, while  preserving content. We focus on Italian in this contribution, but the methodology we propose is obviously applicable to any language for which data is available.

\paragraph{Collection} We used a corpus of around 152K article-headline pairs from two wide circulation Italian newspapers at opposite ends of the political spectrum
namely \textit{la Repubblica} (left-wing) and \textit{Il Giornale} (right-wing) provided by \newcite{invisible-dema-2020}. The data is balanced across the two sources. 
Though we are concerned with headlines, full articles are used in two ways: 
(a) \textit{alignment}; and (b) 
the consistency check classifiers (see  Section~\ref{sec:eval} for details). For the former, we leverage the alignment procedure proposed by \citet{cafagna2019embeddings} and we split our dataset into strongly aligned, weakly aligned and non-aligned news. The purpose of alignment is to control for potential topic biases in the two newspapers so as to better disentangle newspaper-specific style. Additionally, this information is useful in the creation of our datasets, specifically as it addresses the need for parallel data for our evaluation classifiers and the translation-based model (see below).

\paragraph{Alignment}
\label{align}
We compute the tf-idf vectors of all the articles of both newspapers and create subsets of relevant news filtering by date, i.e. considering only news which were published approximately within the same, short time interval
for the two sources. On the tf-idf vectors we then compute cosine similarities for all news in the resulting subset, rank them, and retain only the alignments that are above a certain threshold. The threshold is chosen taking into consideration a trade-off between number of documents and quality of alignment. We choose two different thresholds: one is stricter ($>0.5$) and we use it to select the best alignments; the other one is looser ($>0.185$, and $<=0.5$). 

\paragraph{Data splitting}
We split the dataset into \textit{strongly aligned news}, which are selected using the stricter threshold ($\sim$20K aligned pairs), and \textit{weakly aligned and non-aligned news} ($\sim$100K article-headline pairs equally distributed among the two newspapers). The aligned data is further split as shown in Figure~\ref{fig:data-setting}. \texttt{SA} is left aside and used as test set for the final style transfer task. The remaining three sets are used for training the evaluation classifiers and the models for the target task in various combinations. These are described in Figure~\ref{fig:trainsets} and in connection with the systems' descriptions.\footnote{Note that all sets also always contain the headlines' respective full articles, though these are not necessarily used.}











\section{Systems} 
Our focus is on the interaction of different evaluation settings and approaches to the task.
Accordingly, we develop two different frameworks with different takes on the same problem: (a)
as a true translation task, where given a headline in one style, the model learns to generate a new headline in the target style; (b)
as a summarisation task, where headlines are viewed as an extreme case of summarisation and generated from the article. We exploit article-headline generators trained on opposite sources to do the transfer. This approach does not in principle require parallel data for training. 

For the translation approach (\texttt{S2S}), we train a supervised BiLSTM sequence-to-sequence model with attention from OpenNMT \cite{klein-etal-2017-opennmt}
to map the headline from left-wing to right-wing, and viceversa. Since the model needs parallel data, we exploit the aligned headlines for training. We experiment with three differently composed training sets, varying not only in size, but also in the strength of the alignment, as shown in Figure~\ref{fig:trainsets}.


For the summarisation approach (\texttt{SUM}), we use two pointer-generator networks \cite{see2017get}, which 
include a
\textit{pointing mechanism} able to copy words from the source as well as pick them from a fixed vocabulary, thereby allowing  
better handling of out-of-vocabulary words.
ability to reproduce novel words. 
One model is trained on the \textit{la Repubblica} 
portion of the training set, 
the other on \textit{Il Giornale}.
In a style transfer setting we use these models as follows: Given a headline from \textit{Il Giornale}, for example, the model trained on \textit{la Repubblica} can be run over the corresponding article from \textit{Il Giornale} to generate a headline in the style of \textit{la Repubblica}, and vice versa. To train the models we use subset \texttt{R}, but we also include the lower end of the aligned pairs (\texttt{A3}), see Figure~\ref{fig:trainsets}.

\section{Evaluation}
\label{sec:eval}

Our fully automatic strategy is based on a series of classifiers to assess style strength and content preservation. For style, we train a single classifier (\textit{main}). For content, we train two classifiers that perform two `consitency checks': one ensures that the two headlines (original and transformed) are still compatible (\textit{HH classifier}); the other ensures that the headline is still compatible with the original article (\textit{AH classifier}). See also Figure~\ref{fig:data-setting}.



In what follows we describe these classifiers in more detail. When discussing results, we will show how the contribution of each classifier is crucial towards a comprehensive evaluation.


\paragraph{Main classifier} The main classifier uses a pre-trained BERT encoder with a linear classifier on top fine-tuned with a batch size of 256 and sequences truncated at 32 tokens for 6 epochs with learning rate 1e-05.
Given a headline, this classifier
can distinguish  the two sources
with an f-score of approximately 80\% (see Table~\ref{tab:gold-performance}). Since style transfer is deemed successful if the original style is lost in favour of the target style, we use this classifier to assess how many times a style transfer system manages to reverse the main classifier's decisions. 


\paragraph{HH classifier} This classifier checks compatibility between the original and the generated headline. We use the same architecture as for the main classifier with a slightly different configuration: max. sequence length of 64 tokens, batch size of 128 for 2 epochs (early-stopped), with learning rate 1e-05.
Being trained on strictly aligned data as positive instances (\texttt{A1}), with
a corresponding amount of random pairs as negative instances, it should learn whether two headlines describe the same content or not. 
Performance on gold data is .96 (Table~\ref{tab:gold-performance}).



\paragraph{AH classifier} This classifier performs yet another content-related check. It takes a headline and its corresponding article, and tells whether the headline is appropriate for the article.The classifier is trained on article-headline pairs from both the strongly aligned and the weakly and non-aligned instances (\texttt{R+A3+A1}, Figure~\ref{fig:trainsets}). At test time, the generated headline is checked for compatibility against the source article. We use the same base model as for the main and HH classifiers with batch size of 8, same learning rate and 6 epochs. Performance on gold data is $>$.97 (Table~\ref{tab:gold-performance}).





\begin{table}[h]
\centering
\small
\begin{tabular}{l|l|r|r|r}
\toprule
 \multicolumn{2}{c|}{}  & \textbf{prec} & \textbf{rec} & \textbf{f-score} \\ \midrule
\multirow{2}{*}{main}& \textbf{rep} & 0.77               & 0.83            & 0.80               \\ 
& \textbf{gio} & 0.84               & 0.78            & 0.81               \\ \bottomrule

\multirow{2}{*}{HH} & \textbf{match}    & 0.98               & 0.95            & 0.96             \\ 
& \textbf{no match} & 0.95               & 0.98            & 0.96             \\ \bottomrule

\multirow{2}{*}{AH} & \textbf{match} & 0.96               & 0.99            & 0.98             \\ 
& \textbf{no match} & 0.99               & 0.96            & 0.97             \\ \bottomrule

\end{tabular}
\caption{Performance of the classifiers on gold data.\label{tab:gold-performance}}
\end{table}

\paragraph{Overall compliancy} We calculate a compliancy score which assesses the proportion of times the following three outcomes are successful (i) the \textit{HH classifier} predicts `match'; (ii) the \textit{AH classifier} predicts `match'; (iii) the \textit{main classifier}'s decision is \textit{reversed}. As upperbound, we find the compatibility score for gold at 74.3\% for transfer from \textit{La Repubblica} to \textit{Il Giornale} (\textsl{rep2gio}), and 78.1\% for the opposite direction (\textsl{gio2rep}).


\section{Results and Discussion}

\begin{table*}[!t]
\centering
{\begin{tabular}{ll|rrrrrr}
\toprule
 &  & \textbf{HH} & \textbf{AH} & \textbf{Main} & \textbf{Compl.} & \textbf{BLEU} & \textbf{ROUGE} \\ 

\bottomrule
\rowcolor{lightgray}
\multicolumn{8}{c}{without top aligned data}\\
\toprule

\multirow{ 3}{*}{\texttt{SUM}} & \multicolumn{1}{l|}{\textbf{rep2gio}} & .649               & .876               & .799      & .449  &  .020 & .145\\ 

& \multicolumn{1}{l|}{\textbf{gio2rep}} & .639               & .871               & .435             & .240     &    .026 & .156 \\

& \multicolumn{1}{l|}{\textbf{avg}}           & \bf{.644}               & \bf{.874}                & .616             & .345  &  \bf{.023}  & \bf{.151} \\ \midrule







\multirow{ 3}{*}{\texttt{S2S1}} & \multicolumn{1}{l|}{\textbf{rep2gio}} & .632   & .842      & .815        & .436     & .011  & .136 \\ 
& \multicolumn{1}{l|}{\textbf{gio2rep}} & .444    & .846      & .864   & .321     & .012  & .130     \\ 
& \multicolumn{1}{l|}{\textbf{avg}}     & .538    & .844      & \bf{.840}       & \bf{.379} &    .012     & .133 \\  
\bottomrule
\rowcolor{lightgray}
\multicolumn{8}{c}{with top aligned data}\\
\toprule

\multirow{ 3}{*}{\texttt{S2S2}} & \multicolumn{1}{l|}{\textbf{rep2gio}} & .860   & .845    & .845   & .549  & .018   & .159\\ 
& \multicolumn{1}{l|}{\textbf{gio2rep}} & .612        & .846               & .847              & .442  & .016  &  .151 \\ 
& \multicolumn{1}{l|}{\textbf{avg}}   & .736          & \bf{.846}          & \bf{.849}              & \bf{.496}   & \bf{.017} & \bf{.155} \\ \midrule 

\multirow{ 3}{*}{\texttt{S2S3}} & \multicolumn{1}{l|}{\textbf{rep2gio}} & .728       & .844               & .845        & .520    & .012 &  .139  \\ 
& \multicolumn{1}{l|}{\textbf{gio2rep}} & .760  & .848          & .649             & .420  & .013  & .156\\
& \multicolumn{1}{l|}{\textbf{avg}} & \bf{.744}       & \bf{.846}          & .747        & .470  &  .013 &  .148 \\

\bottomrule
\end{tabular}
}
\caption{Performance on test data.}
\label{tab:test-results}
\end{table*}

Table \ref{tab:test-results} reports results of our evaluation methods both for the summarization system (\texttt{SUM}) and for the style transfer systems (S2S) in the different training set scenarios. 

The top panel in Table \ref{tab:test-results} shows the results for systems where training data is weakly aligned or unaligned. The summarisation system \texttt{SUM} does better at content preservation (HH and AH) than \texttt{S2S1}. However, its scores on the \textit{main} classifier are worse in both transfer directions, as well as on average. The average compliancy score is higher for \texttt{S2S1}. In summary, for data which is not strongly aligned, our methods suggest that style transfer is better when conceived as a translation task. BLEU is higher for \texttt{SUM}, but the overall extremely low scores across the board suggest that it might not be a very informative metric for this setup, although commonly used to assess content preservation in style transfer \cite{rao2018dear}. Our HH and AH classifiers appear more indicative  in this respect, and ROUGE scores seem to correlate a bit more with them, when compared to BLEU. It remains to be investigated whether BLEU, ROUGE, and our content-checking classifiers do in fact measure something similar or not.

With better-aligned data (bottom panel), the picture is more nuanced. Here, the main comparison is between two systems trained on strongly aligned data, one of which (\texttt{S2S2}) has additional, weakly aligned data. The overall compliancy score suggests that this improves style transfer (and this system is also the top performing one over all, also outperforming \texttt{S2S1} and \texttt{SUM}). As for content preservation (AH and HH scores), \texttt{S2S3} is marginally better on average for HH, but not for AH, where the two systems are tied. 

Overall, the results of the classification-based evaluation also highlight a difference between a summarisation-based system (\texttt{SUM}), which tends to be better at content preservation, compared to a translation-based style transfer setup (especially \texttt{S2S2}) which transfers style better. Clearly, a corpus-based metric such as BLEU fails to capture these distinctions, but here does not appear informative even just for assessing content preservation.

One aspect that will require further investigation, since we do not have a clear explanation for it as of now, is the performance difference between the two translation directions. Indeed, transforming a \textit{La Repubblica} headline into a \textit{Il Giornale} headline appears more difficult than transforming headlines in the opposite directions, under most settings.

\section{Conclusions}
This paper addressed the issue of how to evaluate style transfer. We explicitly compared systems in terms of the extent to which they preserve content, and their success at transferring style. The latter is known to be hard for humans to evaluate \cite{dethlefs2014cluster,invisible-dema-2020}. Our aim was primarily to see to what extent different evaluation strategies based on purposely trained classifiers could distinguish between models, insofar as they perform better at either of these tasks and in different training scenarios. 

Our findings suggest that our proposed combination of classifiers focused on both content and style transfer can potentially help to distinguish models in terms of their strengths. Interestingly, a commonly used metric such as BLEU does not seem to be informative in our experiments, not even for the content preservation aspects. 

To the extent that stylistic distinctions remain hard for humans to evaluate in setups such as the one used here, a classification-based approach with consistency checks for content preservation is a promising way forward, especially to support development in a relatively cheap and effective way. 

Future work will have to determine how the various metrics we have used relate to each other (especially our classifiers and BLEU/ROUGE), and whether human judgement can be successfully brought back, and in case in what form, at some stage of the evaluation process.







\bigskip

\bibliography{anthology,acl2020,headlines}
\bibliographystyle{acl_natbib}

\end{document}